\documentclass[preprint,authoryear,12pt]{elsarticle}
\usepackage{amsmath,amssymb,amsfonts}
\usepackage{mathtools}
\usepackage{bm}
\usepackage{siunitx}

\usepackage{booktabs}
\usepackage{array}
\usepackage{tabularx}
\usepackage{multirow}
\usepackage{ragged2e}
\newcolumntype{L}{>{\RaggedRight\arraybackslash}X}
\usepackage{adjustbox}

\usepackage{subcaption}
\usepackage{enumitem}

\usepackage{tikz}
\usetikzlibrary{positioning,arrows.meta,fit,shapes.geometric,calc}

\usepackage[hyphens]{url}

\biboptions{sort&compress}
\journal{Engineering Applications of Artificial Intelligence}

\makeatletter
\long\def\panel@gobble#1{}
\long\def\panel@gobbleopt[#1]#2{}
\long\def\panel@caption{\@ifnextchar[{\panel@gobbleopt}{\panel@gobble}}
\newcommand{\panelinput}[1]{%
  \begingroup
  \renewenvironment{figure}[1][]{}{}%
  \renewenvironment{table}[1][]{}{}%
  \let\caption\panel@caption
  \let\label\panel@gobble
  \input{#1}%
  \endgroup}
\makeatother

\begin{document}

\begin{frontmatter}

% EAAI desk-rejection condition: "The use of undefined acronyms in the title and
% in the abstract is forbidden." "Surface-EMG" expanded accordingly.
\title{Personalising a Cross-User Surface Electromyography Encoder Under a Small
Calibration Budget}

\author[bme]{Jethro Odeyemi}
\ead{jethro.odeyemi@usask.ca}

\author[bme]{W.J. (Chris) Zhang\corref{cor1}}
\ead{chris.zhang@usask.ca}
\cortext[cor1]{Corresponding author.}

\affiliation[bme]{organization={Division of Biomedical Engineering, University of Saskatchewan},
            addressline={57 Campus Drive},
            city={Saskatoon},
            state={SK},
            postcode={S7N 5A9},
            country={Canada}}

\begin{abstract}
A myoelectric interface needs calibration from the user before it will function. Earlier work
has treated calibration as a quantity, but has not asked the question of what a device should
do with the calibration repetitions once they have been collected. This paper views
personalizing the cross-user encoder as a design decision with a cost. Four alternative
approaches to using exactly the same labeled repetitions were tested from a single cross-user
encoder per held-out subject. Prototypical adaptation, linear probes, scaled fine-tuning and
full fine-tuning were tested at every budget up to the maximum each database allows, five
repetitions on DB1 and four on DB2 and DB5. Comparing four ways to spend a small calibration
budget across 77 subjects, full fine-tuning is the most accurate at every budget, consistently
enough that there is no exception among subsets of subjects. The result which impacts how one
might make an engineering decision however is that a gradient free prototypical rule recovers
52 to 78 per cent of its benefit with no optimiser and no per-user copy of the weights, which
makes personalisation something a worn device can do at donning time. The widespread intuition
that a good representation only needs a fresh classifier is incorrect here. How well each
method may perform relative to a per-user classifier that would be fitted by a clinic will
depend on the specific database.
\end{abstract}


\end{frontmatter}

\section{Introduction}

A myoelectric interface needs calibration from the user before it will function. In earlier
work~\citep{odeyemi_encoder} we built an encoder that was able to decode a user it had never
previously seen without any calibration at all but could only decode far too little to be
useful. We also demonstrated that a few labeled repetitions would take us past a per-user linear
classifier on DB1 and DB2, although we trailed on the ten-subject DB5. Similarly, a companion
study~\citep{odeyemi_intersession} showed that a single labeled repetition taken after a first
session would recover approximately as much as the best label free adaptation. While both of
these studies have treated calibration as a quantity, neither asked the question of what a
device should do with the calibration repetitions once they have been collected.

This question is important because there is obviously an expense associated with taking the next
step and doing something with those repetitions. The obvious path is to fine tune the entire
network on those repetitions; and fine tuning does produce good results. However, fine tuning
requires an optimizer to be run either on the device or in the cloud; and it also creates the
requirement for a separate model to be stored for every user; and the process takes time as it
converges. Therefore, if there is a cheaper way to use the same amount of labeling that gets
close, or is ahead when the labels are very few, that changes the overall engineering of the
product and not just the numbers in the tables.

This paper views personalizing the cross-user encoder as a design decision with a cost. Four
alternative approaches to using exactly the same labeled repetitions were tested from a single
cross-user encoder per held-out subject. Prototypical adaptation, linear probes, scaled
fine-tuning and full fine-tuning were tested at every budget up to the maximum each database
allows, five repetitions on DB1 and four on DB2 and DB5. All four approaches adapt the same base
model using the same labeled repetitions from each held-out subject. The base model is trained
on the remaining subjects based on the same leave-one-subject-out protocol presented in
Section~\ref{sec:ch3-protocol}. Every approach uses the same base model, the same labeled
repetitions and the same training budget. As such, the full-fine-tuned arm represents an
independent repetition of the calibration procedure presented in earlier
work~\citep{odeyemi_encoder} rather than the same run reported twice. Additionally, the
full-fine-tuning arm is within a few thousandths of that study's calibration procedure on DB1
and about a point on DB2 thereby establishing a fixed run-to-run variability and providing a
reference point for the other three alternatives.

\section{Background and positioning}

Personalizing a generic model to represent a particular user is essentially identical to
adapting a pre-trained network to a small labeled task. The machine learning community has
mapped this space well. One extreme includes methods known as metric learning wherein each class
is represented by the average of its embeddings and a query is matched to the class whose
prototype embedding is closest to that query's embedding~\citep{snell2017prototypical}. Since no
gradients are generated during this process, it only requires one forward pass through the
network for each labeled window and is stable regardless of whether it is given hundreds of
thousands of examples or merely dozens. At the opposite end of this spectrum lies fine-tuning.
Fine-tuning updates all of a network's own weights allowing it to potentially reshape
representations for new users. The downside of this is that fine-tuning typically involves
running an optimization routine and creating a separate copy of the model for each user.
Somewhere between these two extremes lie methods referred to as linear probing wherein the final
classifier is refitted to match new data while leaving the learned representation
unchanged~\citep{kornblith2019better}. There are numerous partial update methods that adjust
only a subset of learnable parameters or hold certain parts of the network back by reducing
their corresponding learning rates~\citep{howard2018ulmfit}.

Surface EMG research has employed this spectrum somewhat unevenly. Fine-tuning a network on
labeled data from new users is widely practiced in surface EMG research for applications
involving myoelectric transfer learning~\citep{coteallard2019deep,ketyko2019domain}; and
similarly throughout much of transfer learning
literature~\citep{coteallard2019deep,coteallard2020interpreting,lobov2018latent,shen2022generalization,alfaro2022user}.
Furthermore, our systematic review~\citep{odeyemi_review} identified twenty-two out of 1077
studies that report experimental results with an explicitly defined few-shot calibration budget.
These studies are included in our review
corpus~\citep{chen2020handa,ameri2019deep,tam2021intuitive,kobylarz2020thumbs,shi2023physics,nguyen2024frequency}.
Prototype and/or metric-based recognition methods have appeared in surface EMG research
primarily to enable addition of new gestures rather than to personalize an existing gestural
vocabulary. The field has not yet established a like-for-like comparison: i.e., the same
encoder, the same held-out subjects, the same labeled repetitions, multiple ways to spend them;
along with reporting of each approach's cost in terms of computational resources required; and
comparative reporting of each approach's performance.

The encoder personalised here is the montage-agnostic cross-user encoder of our earlier
work~\citep{odeyemi_encoder}, which also established the account of when such an encoder exceeds
a per-user baseline and when it does not. There is no single answer among the three datasets
studied. The encoder outperforms the per-user LDA baseline substantially on DB1, slightly on DB2
after being trained adequately, and falls behind it on DB5. Notably, none of these outcomes
depend on sampling rate or number of channels directly and not even on number of training
subjects over the range available. What tracks that ordering is the strength of the per-user LDA
baseline. As signal quality improves (measured through fidelity), so too does performance of the
per-user LDA baseline: on DB1, it achieves an average macro-F1 value of $0.593$ at 100~Hz; on
DB5, it achieves an average macro-F1 value of $0.802$ at 200~Hz; and on DB2 it achieves an
average macro-F1 value of $0.857$ at 2~kHz. Hand-crafted features obviously have greater
bandwidth to capitalize upon as sampling rates increase. Thus, the margins open to trainable
encoders are least when baselines are strongest. Separately, constraints placed by limiting
subject diversity are a stability bound rather than a continuous decline. Down to nine subjects
however, the encoder retains most of its advantages versus the per-user classifier; below that it
ceases training altogether as opposed to degrading progressively.

\section{Method}

\subsection{Montage-agnostic encoder architecture}
\label{sec:ch3-method}

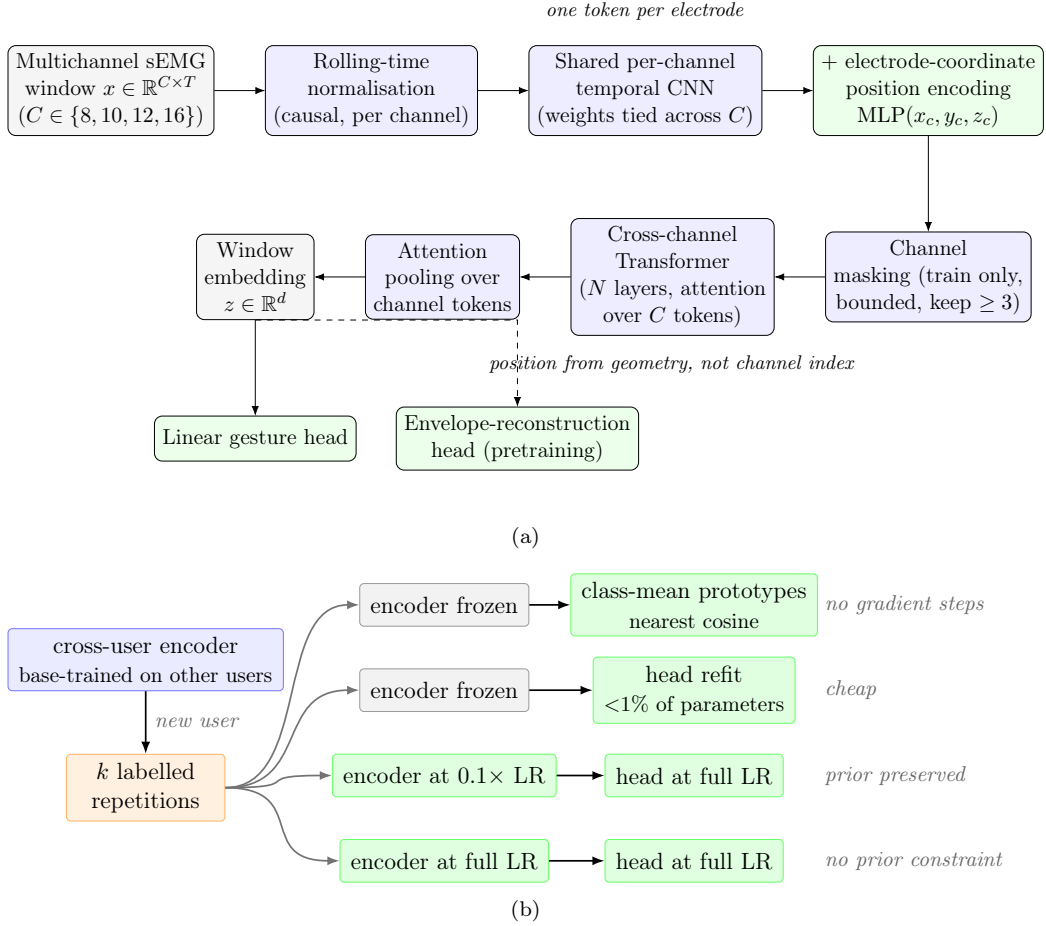
\begin{figure}[tp]\centering
\begin{subfigure}{\linewidth}\centering
% Fused montage-agnostic EMG encoder (Chapter 3). Author's own architecture.
\begin{figure}[t]
\centering
\resizebox{\textwidth}{!}{%
\begin{tikzpicture}[
  font=\small,
  box/.style={draw, rounded corners, align=center, minimum height=8mm, inner sep=4pt, fill=black!4},
  op/.style={draw, rounded corners, align=center, minimum height=8mm, inner sep=4pt, fill=blue!7},
  hl/.style={draw, rounded corners, align=center, minimum height=8mm, inner sep=4pt, fill=green!8},
  ->, >=Latex, node distance=6mm and 9mm]

\node[box] (in) {Multichannel sEMG\\window $x\in\mathbb{R}^{C\times T}$\\($C\in\{8,10,12,16\}$)};
\node[op, right=of in] (rtn) {Rolling-time\\normalisation\\(causal, per channel)};
\node[op, right=of rtn] (cnn) {Shared per-channel\\temporal CNN\\(weights tied across $C$)};
\node[hl, right=of cnn] (pos) {$+$ electrode-coordinate\\position encoding\\$\mathrm{MLP}(x_c,y_c,z_c)$};

\node[op, below=17mm of pos] (acm) {Channel\\masking (train only,\\bounded, keep $\ge 3$)};
\node[op, left=of acm] (tf) {Cross-channel\\Transformer\\($N$ layers, attention\\over $C$ tokens)};
\node[op, left=of tf] (pool) {Attention\\pooling over\\channel tokens};
\node[box, left=of pool] (out) {Window\\embedding\\$z\in\mathbb{R}^{d}$};

\draw (in) -- (rtn);
\draw (rtn) -- (cnn);
\draw (cnn) -- (pos);
\draw (pos.south) -- (acm.north);
\draw (acm) -- (tf);
\draw (tf) -- (pool);
\draw (pool) -- (out);

\node[align=center, font=\footnotesize, above=3mm of cnn] {\emph{one token per electrode}};
\node[align=center, font=\footnotesize, fill=white, inner sep=1pt, below=3mm of tf]
  {\emph{position from geometry, not channel index}};

% heads
\node[hl, below=17mm of out] (head) {Linear gesture head};
\node[hl, right=7mm of head] (mae) {Envelope-reconstruction\\head (pretraining)};
\draw (out.south) -- (head.north);
\draw[dashed] (out.south) -| (mae.north);
\end{tikzpicture}%
}
\caption[Fused montage-agnostic encoder]{The proposed montage-agnostic encoder. Each electrode
becomes a single token whose position is supplied by its forearm coordinate rather than its channel
index, so one set of weights ingests any electrode count. Rolling-time normalisation makes the
representation calibration-free, cross-channel attention mixes the electrode tokens, and bounded
channel masking regularises toward user-invariant features. The linear head is used for gesture
classification; the dashed head is the self-supervised pretraining branch.}
\label{fig:ch3-encoder}
\end{figure}
\caption{}\label{fig:ch3-encoder}
\end{subfigure}

\vspace{1.5ex}

\begin{subfigure}{\linewidth}\centering
\resizebox{\linewidth}{!}{\begin{figure}[t]\centering
\begin{tikzpicture}[
  font=\small,
  b/.style={rounded corners=2pt, draw, minimum height=7mm, inner xsep=5pt, align=center},
  enc/.style={b, fill=blue!8, draw=blue!55},
  frz/.style={b, fill=black!5, draw=black!40},
  upd/.style={b, fill=green!12, draw=green!60},
  cal/.style={b, fill=orange!12, draw=orange!60},
  arr/.style={-{Latex[length=2mm]}, thick},
  lbl/.style={font=\footnotesize\itshape, text=black!60}]

% shared base
\node[cal, minimum width=26mm] (cal) at (0,0) {$k$ labelled\\ repetitions};
\node[enc, minimum width=30mm] (base) at (0,2.1) {cross-user encoder\\ \footnotesize base-trained on other users};
\draw[arr] (base.south) -- node[lbl,right]{new user} (cal.north);

% strategy 1: prototypical (no gradient)
\node[frz, minimum width=22mm] (p1) at (4.9,3.0) {encoder frozen};
\node[upd, minimum width=26mm] (p2) at (9.0,3.0) {class-mean prototypes\\ \footnotesize nearest cosine};
\draw[arr] (p1) -- (p2);
\node[lbl, anchor=west] at (11.0,3.0) {no gradient steps};

% strategy 2: linear probe
\node[frz, minimum width=22mm] (h1) at (4.9,1.6) {encoder frozen};
\node[upd, minimum width=26mm] (h2) at (9.0,1.6) {head refit\\ \footnotesize ${<}1\%$ of parameters};
\draw[arr] (h1) -- (h2);
\node[lbl, anchor=west] at (11.0,1.6) {cheap};

% strategy 3: scaled fine-tune
\node[upd, minimum width=22mm] (e1) at (4.9,0.2) {encoder at $0.1\times$ LR};
\node[upd, minimum width=26mm] (e2) at (9.0,0.2) {head at full LR};
\draw[arr] (e1) -- (e2);
\node[lbl, anchor=west] at (11.0,0.2) {prior preserved};

% strategy 4: full fine-tune
\node[upd, minimum width=22mm] (f1) at (4.9,-1.2) {encoder at full LR};
\node[upd, minimum width=26mm] (f2) at (9.0,-1.2) {head at full LR};
\draw[arr] (f1) -- (f2);
\node[lbl, anchor=west] at (11.0,-1.2) {no prior constraint};

% fan out from calibration set
\draw[arr, black!55] (cal.east) .. controls +(1.4,0) and +(-1.4,0) .. (p1.west);
\draw[arr, black!55] (cal.east) .. controls +(1.4,0) and +(-1.4,0) .. (h1.west);
\draw[arr, black!55] (cal.east) .. controls +(1.4,0) and +(-1.4,0) .. (e1.west);
\draw[arr, black!55] (cal.east) .. controls +(1.4,0) and +(-1.4,0) .. (f1.west);

\end{tikzpicture}
\caption{The four adaptation strategies. All start from the same cross-user
encoder and the same $k$ labelled repetitions from the new user. Prototypical adaptation changes
no weights at all, the linear probe moves only the head, and the two fine-tuning variants move
the whole model, differing in whether the encoder is held back by a reduced learning rate.}
\label{fig:ch5-strategies}
\end{figure}}
\caption{}\label{fig:ch5-strategies}
\end{subfigure}
\caption{(a) The proposed montage-agnostic encoder. Each electrode
becomes a single token whose position is supplied by its forearm coordinate rather than its channel
index, so one set of weights ingests any electrode count. Rolling-time normalisation makes the
representation calibration-free, cross-channel attention mixes the electrode tokens, and bounded
channel masking regularises toward user-invariant features. The linear head is used for gesture
classification; the dashed head is the self-supervised pretraining branch.
(b) The four adaptation strategies. All start from the same cross-user
encoder and the same $k$ labelled repetitions from the new user. Prototypical adaptation changes
no weights at all, the linear probe moves only the head, and the two fine-tuning variants move
the whole model, differing in whether the encoder is held back by a reduced learning rate.}
\label{fig:ch5-method}
\end{figure}

We begin by providing details concerning our encoder's design. The overall architecture is
depicted in Figure~\ref{fig:ch5-method}(a). Our encoder maps sequences of EMG signals ($x \in
\mathbb{R}^{C\times T}$) onto fixed-size embeddings ($z$), where $C$ represents the possible
number of channels in any of the available montages.

Our proposed encoder is founded on the idea that each electrode is treated as one independent
token whose identity corresponds to its physical position on the forearm.

% Consolidated method schematic: the Chapter-3 encoder diagram and the
% four-strategy diagram, previously two separate floats, as panels (a) and (b).

\paragraph{Causal Rolling-Time normalisation} Our first layer normalizes each channel
individually using their respective causal rolling statistics over time; these statistics
represent expanding window means and variances calculated over time. Since our normalizing is
done per channel and only uses causal information from past time steps, at runtime a previously
unobserved user will self-normalize their input data with respect to their own signal using only
their own causal history; no user-specific statistics will be transported from training to
testing. Hence, even though there is no labeled data from the new user available at runtime, our
model will still adaptively scale its input representations according to new users' signals
without requiring any additional labels.

\paragraph{Shared per-channel tokeniser} To produce one feature vector per electrode, we apply
a single one-dimensional convolutional stack separately to every channel. Due to sharing of
weights across channels, adding or removing an electrode simply adds/removes a token to/from our
tokenizer without modifying any parameter(s).

\paragraph{Electrode-coordinate position encoding} Each electrode contains normalized $(x,y,z)$
coordinates specifying its position on the forearm. These coordinates are mapped to vectors
added to each electrode token by means of a multilayer perceptron. Therefore, our model learns
where each electrode resides, not merely its index, thus allowing any electrode configuration to
map into one uniform representation space, with each electrode serving as an unordered element
within that space.

\paragraph{Cross-channel attention} Our transformer encoder~\citep{vaswani2017attention}
applies attention mechanisms across our electrode tokens. The temporal structure is encoded via
our per-channel tokenizers; attention mechanisms capture spatial relations between electrodes.
Following application of attention mechanisms across all channels, an attention-pooling layer is
used to collapse the electrode tokens into a single window embedding; finally a linear head
computes gesture logits.

\paragraph{Channel-masking regularisation} During training time, a bounded number of electrodes
is randomly masked such that at least three channels remain active for each montage. This forces
our model towards features that are invariant or robust with regards to missing or displaced
electrodes rather than relying on any single channel. How many electrodes should be masked
affects whether masking fails completely: if too aggressive on very small montages (e.g., all
electrodes masked), there are no valid windows containing active electrodes resulting in
collapsed training; we solved this issue by enforcing bounds for masking.

\paragraph{Implementation} The encoder has $d{=}256$, four transformer layers, and
approximately 3.3M total parameters. Training used the AdamW
optimizer~\citep{loshchilov2019decoupled} along with a one cycle schedule.

\subsection{Calibration-light evaluation protocol}
\label{sec:ch3-protocol}

All evaluations are conducted cross-user (leave-one-subject-out). Training occurs exclusively on
all subjects excluding one; this held-out subject is divided into calibration repetitions and
test repetitions. The model is updated/calibrated on the calibration repetitions and evaluated
on disjoint test repetitions. At each calibration budget, predictions are made using majority
vote per-repetition.

No data from the held-out subject is ever seen by our training code or model selection code at
any calibration budget; exactly the same windowing and evaluation criteria are applied equally
to our model as applied to all baseline methods.

Primary metrics reported for each method are macro-average F1-score (macro-$F_1$) with
associated distributions across subjects. Macro-$F_1$ is selected instead of pure accuracy due
to imbalance between gesture classes as well as between support counts per-gesture class; hence,
accuracy would misrepresent models favoring dominant classes.

\subsection{Shared base model}

Each adaptation strategy begins with the same base model. Specifically, for each held-out
subject, the shared montage-agnostic encoder described in Section~\ref{sec:ch3-method} is
trained on all remaining subjects using the exact same leave-one-subject-out protocol of
Section~\ref{sec:ch3-protocol} with exactly the same hyper-parameters. Once this shared base
model has been trained, copies are made for each adaptation strategy to apply adaptations using
the same $k$ labeled repetitions from each held-out subject. Thus, any differences among results
are due solely to adaptations applied to their respective copies and not due to differing
starting points.

\subsection{Adaptation strategies}

Figure~\ref{fig:ch5-method}(b) describes all four strategies.

\emph{Prototypical adaptation}: Embeddings are created using the frozen encoder for each labeled
repetition. The average embedding for each gesture (class) forms a class prototype. Each test
window is classified via cosine similarity to its most similar prototype. No gradients are
generated nor are any weights updated. Gestures never shown by the calibration set have no
prototype and therefore cannot be predicted.

\emph{Linear probe}: The encoder is frozen; and the classifier head is refitted using the same
labeled repetitions as calibration windows. Fewer than one percent of all weights are modified
since only the final classification layer is being adjusted.

\emph{Scaled fine-tune}: The entire model is trained using the calibration windows; however,
scaling down by an order-of-magnitude the learning rate of the encoder relative to the head
allows retaining prior knowledge embodied in cross-user representations while simultaneously
adapting to new user-specific representations.

\emph{Full fine-tune}: The entire model is trained using a single learning rate. This is
identical to our earlier calibration procedure~\citep{odeyemi_encoder} and serves as our primary
reference point.

\subsection{Protocol and metrics}

Our design protocol follows identically the leave-one-subject-out design protocol of
Section~\ref{sec:ch3-protocol}. Calibration repetitions come from each subject's designated
calibration repetitions. Scoring always comes from each subject's designated testing repetitions
which none of our strategies will see. Budgets of one, two, three, and five repetitions will be
tested on DB1; whereas DB2 and DB5 designate only four possible calibration repetitions, two of
their six repetitions being reserved for testing purposes. Thus, their highest possible budget
will also be four. Macro F1 scored at trial level by voting by majority over windows per gesture
repetition and averaged across classes, represents our primary metric for evaluation; it is used
throughout this work. Reported statistics will be averages and standard deviations across
held-out subjects; along with per-subject statistics so that variability between sub-scores can
be viewed for each mean.

In addition to accuracy; we will report on each strategy's cost for adapting: (a) Number of
gradient steps taken; and (b) Fraction of parameters modified by each strategy. These two
statistics are necessary in order to accurately compare strategies that differ by orders of
magnitude in what they require from a device.

\section{Experimental setup}

% Consolidated experimental-setup table: the database table and the
% hyper-parameter table, previously two separate floats, as (a) and (b).
\begin{table}[tp]\centering
\caption{Experimental setup. (a) Databases used for the personalisation study. All three are evaluated leave-one-subject-out with the same calibration protocol. (b) Training and calibration configuration for the personalisation experiments.}
\label{tbl:ch5-setup}
\begin{subtable}{\linewidth}\centering
\caption{}\label{tbl:ch5-datasets}
\begin{table}[t]\centering
\caption{Databases used for the personalisation study. All three are evaluated leave-one-subject-out with the same calibration protocol.}\label{tbl:ch5-datasets}
\begin{tabular}{lcccc}
\toprule
Database & Subjects & Gestures & Electrodes & Sampling rate \\
\midrule
DB1 & 27 & 52 + rest & 10 & 100 Hz \\
DB2 & 40 & 49 + rest & 12 & 2 kHz \\
DB5 & 10 & 41 + rest & 16 & 200 Hz \\
\bottomrule
\end{tabular}
\end{table}

\end{subtable}

\vspace{1.5ex}

\begin{subtable}{\linewidth}\centering
\caption{}\label{tbl:ch5-hyperparams}
\begin{table}[t]\centering
\caption{Training and calibration configuration for the personalisation experiments.}\label{tbl:ch5-hyperparams}
\begin{tabular}{ll}
\toprule
Hyper-parameter & Value \\
\midrule
Base-training epochs & 50 \\
Calibration epochs & 30 \\
Base learning rate & $0.0004$ \\
Calibration learning rate & $0.0003$ \\
Encoder LR scale (scaled fine-tune) & 0.1 \\
Channel masking & 0.25 \\
Batch size & 256 \\
Model width $d$ & 256 \\
Transformer layers / heads & 4 / 4 \\
Prototype metric & cosine to class-mean embedding \\
\bottomrule
\end{tabular}
\end{table}

\end{subtable}
\end{table}

Three NinaPro databases~\citep{atzori2014electromyography,pizzolato2017comparison}, listed in
Table~\ref{tbl:ch5-setup}(a), were selected in order to cover various conditions that must be
accounted for by our encoder: DB1 has 27 subjects and a relatively low sampling frequency; DB2
contains 40 subjects sampled at two kHz; and DB5 has 10 subjects with a dual-ring montage. Their
varying number of training subjects and signal fidelity have implications since earlier
work~\citep{odeyemi_encoder} demonstrated that signal fidelity defines strength of a per-user
linear classifier and therefore what the encoder is measured against. Conversely, the number of
training subjects matters mainly as a floor beneath which cross-user training fails completely.
Hyperparameter settings utilized during training are detailed in Table~\ref{tbl:ch5-setup}(b)
and follow identically the encoder study~\citep{odeyemi_encoder} so that our full-fine-tune arm
replicates that study's calibration results.

The comparison baseline for this experiment is a per-user Hudgins and LDA pipeline fit on the
same calibration repetitions, the same baseline used in the encoder
study~\citep{odeyemi_encoder}, which is what a clinic fits for an individual user.

\section{Results}

% Consolidated results table: the headline table and the full strategy-by-budget
% matrix, previously two separate floats, as (a) and (b).
\begin{table}[tp]\centering
\caption{(a) Personalisation headline. Cross-user macro-F1 of each adaptation strategy at the smallest and largest calibration budget, with the unadapted encoder and the per-user LDA baseline for reference. (b) Full personalisation matrix: cross-user macro-F1 for every strategy at every calibration budget, mean over held-out subjects. DB2 and DB5 designate only four calibration repetitions, two of their six being held out for test, so their five-repetition cell does not exist and is dashed.}
\label{tbl:ch5-results}
\begin{subtable}{\linewidth}\centering
\caption{}\label{tbl:ch5-headline}
\begin{table}[t]\centering
\caption{Personalisation headline. Cross-user macro-F1 of each adaptation strategy at the smallest and largest calibration budget, with the unadapted encoder and the per-user LDA baseline for reference.}\label{tbl:ch5-headline}
{\small\setlength{\tabcolsep}{4pt}%
\begin{tabular}{lcccccc}
\toprule
Dataset & Budget & No adapt. & Proto. & Linear probe & Scaled FT & Full FT \\
\midrule
DB1 & 1-shot & 0.103 & 0.428 & 0.169 & 0.380 & 0.523 \\
DB1 & 5-shot & 0.103 & 0.734 & 0.651 & 0.852 & 0.911 \\
DB2 & 1-shot & 0.198 & 0.576 & 0.368 & 0.707 & 0.796 \\
DB2 & 4-shot & 0.198 & 0.733 & 0.772 & 0.953 & 0.981 \\
DB5 & 1-shot & 0.154 & 0.261 & 0.194 & 0.267 & 0.325 \\
DB5 & 4-shot & 0.154 & 0.410 & 0.289 & 0.536 & 0.642 \\
\bottomrule
\end{tabular}}
\end{table}

\end{subtable}

\vspace{1.5ex}

\begin{subtable}{\linewidth}\centering
\caption{}\label{tbl:ch5-matrix}
{\small\begin{table}[t]\centering
\caption{Full personalisation matrix: cross-user macro-F1 for every strategy at every calibration budget, mean over held-out subjects. DB2 and DB5 designate only four calibration repetitions, two of their six being held out for test, so their five-repetition cell does not exist and is dashed.}\label{tbl:ch5-matrix}
\begin{tabular}{llccccc}
\toprule
Dataset & Strategy & 1-shot & 2-shot & 3-shot & 4-shot & 5-shot \\
\midrule
DB1 & prototypical & 0.428 & 0.602 & 0.671 & -- & 0.734 \\
DB1 & linear probe & 0.169 & 0.288 & 0.421 & -- & 0.651 \\
DB1 & scaled fine-tune & 0.380 & 0.617 & 0.741 & -- & 0.852 \\
DB1 & full fine-tune & 0.523 & 0.749 & 0.830 & -- & 0.911 \\
\midrule
DB2 & prototypical & 0.576 & 0.684 & 0.721 & 0.733 & -- \\
DB2 & linear probe & 0.368 & 0.551 & 0.694 & 0.772 & -- \\
DB2 & scaled fine-tune & 0.707 & 0.871 & 0.927 & 0.953 & -- \\
DB2 & full fine-tune & 0.796 & 0.932 & 0.967 & 0.981 & -- \\
\midrule
DB5 & prototypical & 0.261 & 0.352 & 0.378 & 0.410 & -- \\
DB5 & linear probe & 0.194 & 0.230 & 0.253 & 0.289 & -- \\
DB5 & scaled fine-tune & 0.267 & 0.397 & 0.490 & 0.536 & -- \\
DB5 & full fine-tune & 0.325 & 0.487 & 0.588 & 0.642 & -- \\
\bottomrule
\end{tabular}
\end{table}
}
\end{subtable}
\end{table}

\subsection{Strategy ranking by calibration budget}

\begin{figure}[!htbp]\centering
\begin{subfigure}{\linewidth}\centering
\includegraphics[width=0.95\linewidth]{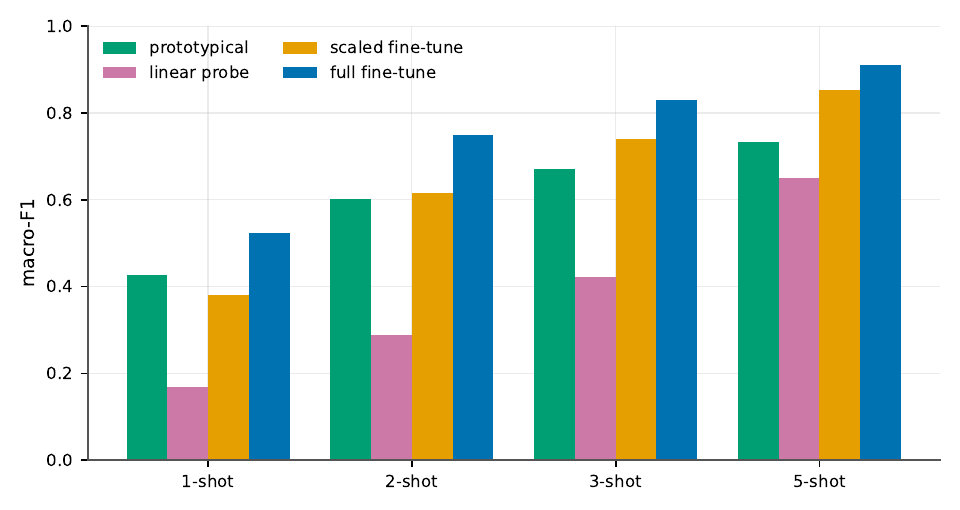}
\caption{}\label{fig:ch5-ranking}
\end{subfigure}

\vspace{1.5ex}

\begin{subfigure}{\linewidth}\centering
\includegraphics[width=0.95\linewidth]{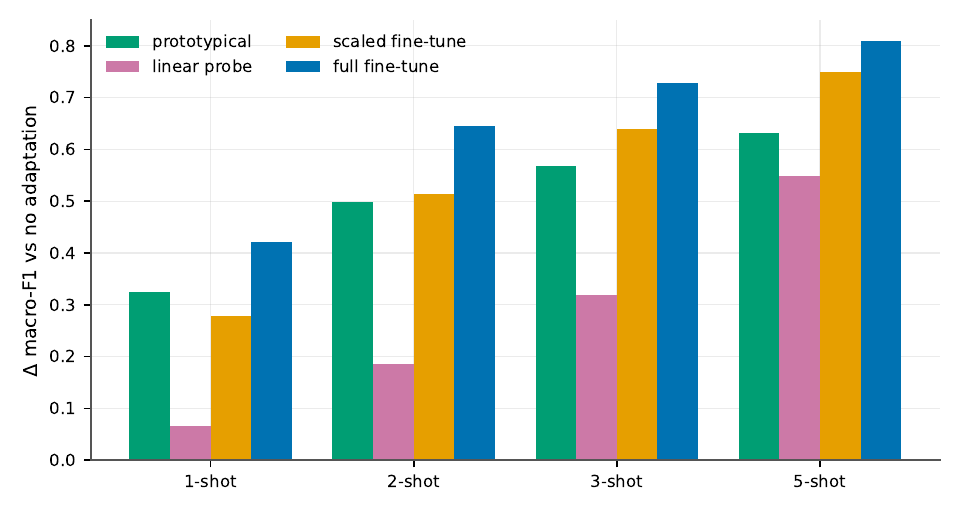}
\caption{}\label{fig:ch5-gain}
\end{subfigure}
\caption{(a) Strategy ranking at each calibration budget on DB1. Full fine-tuning leads at every budget, including the smallest; prototypical adaptation is second at one repetition, ahead of the scaled fine-tune (0.428 against 0.380), and falls behind it from two repetitions on. (b) Gain over the unadapted encoder for each strategy and budget. Every strategy improves on no adaptation, but by different amounts per labelled repetition.}
\label{fig:ch5-ranking-gain}
\end{figure}

At every calibration budget on all three databases full fine-tuning is always the most accurate
strategy (Table~\ref{tbl:ch5-results}(a), Figure~\ref{fig:ch5-ranking-gain}(a)).
Against prototypical adaptation full fine-tuning performs better on all twenty-seven
subjects for the DB1 database and all forty subjects for the DB2 database, and on nine or ten of
the ten subjects for the DB5 database, with paired significance below $10^{-2}$ everywhere and
below $10^{-7}$ for the two larger databases (see Table~\ref{tbl:ch5-comparisons}(b)).

\begin{figure}[p]\centering
\includegraphics[width=0.93\linewidth]{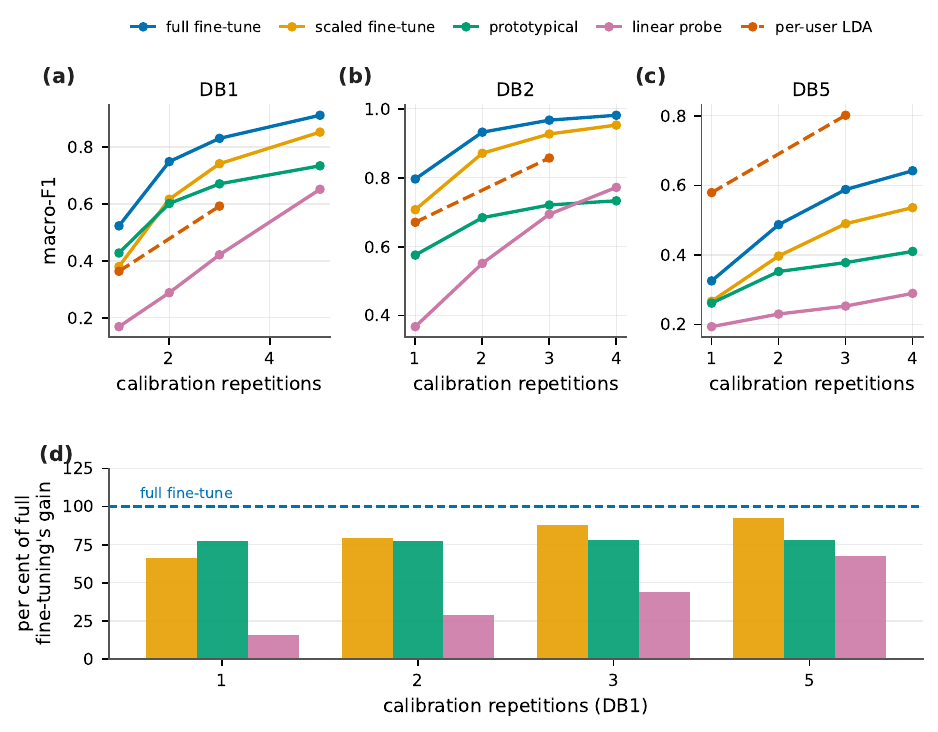}

\vspace{1.5ex}

% the four-panel plot above carries its own baked-in (a)-(d); the heatmap
% continues that lettering as (e) (the sub-counter is reset by \begin{subfigure},
% so the printed letter has to be set directly).
\renewcommand{\thesubfigure}{e}
\begin{subfigure}{\linewidth}\centering
\includegraphics[width=0.95\linewidth]{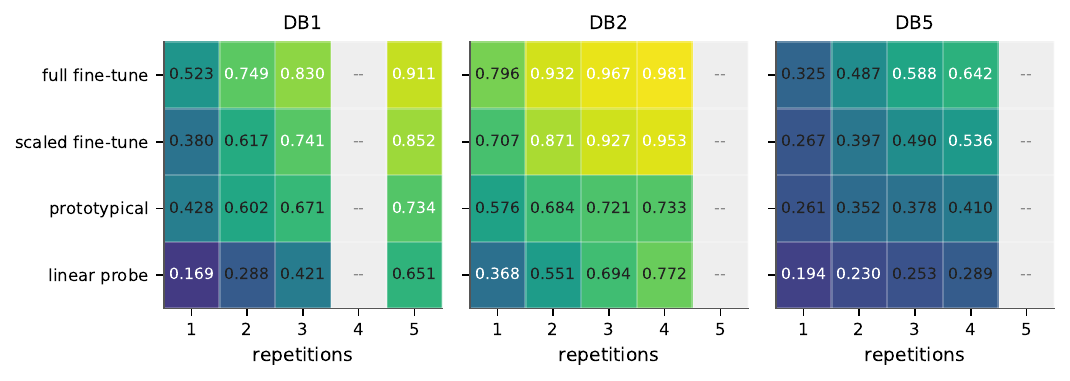}
\caption{}\label{fig:ch5-matrix-heatmap}
\end{subfigure}
\caption{Personalisation strategies under a calibration budget. (a-c) Cross-user macro-F1 against budget for each strategy on DB1, DB2 and DB5, with the per-user LDA baseline dashed. (d) On DB1, the percentage of full fine-tuning's gain over the unadapted encoder that each cheaper strategy recovers at each budget. (e) Full personalisation matrix: cross-user macro-F1 for every strategy at every calibration budget, mean over held-out subjects. Grey cells are budgets a database does not provide, since DB2 and DB5 designate only four calibration repetitions and DB1 five.}
\label{fig:ch5-strategy-panel}
\end{figure}

The performance of the strategies ranked below full fine-tuning can vary depending on the
database and the number of times labelled examples were repeated during the calibration process.
At every calibration budget on the DB5 database, and at every budget up to three labelled
repetitions on the DB2 database, the scaled fine-tune was second followed by prototypical
adaptation third, and the linear probe last. However, at the highest calibration budget on the
DB2 database the linear probe overtook prototypical adaptation (0.772 vs. 0.733). On the DB1
database at the lowest calibration budget (one labelled example), prototypical adaptation
performed better than the scaled fine-tune. Specifically, at one labelled repetition
prototypical adaptation reached a macro F1 value of 0.428 versus 0.380 for the scaled fine-tune.
This indicates that with just one labelled example available prototypical adaptation produces a
stable estimate of a class mean whereas the conservative update used in the scaled fine-tune was
largely noise. This can be seen clearly in the calibration plots shown in
Figures~\ref{fig:ch5-strategy-panel}a-c.

% Consolidated: the four-panel strategy figure keeps its own internal (a)-(d)
% lettering, and the personalisation-matrix heatmap joins it as panel (e).

% Consolidated: the DB1 strategy-ranking bars and the gain-over-no-adaptation
% bars, previously two separate floats, as (a) and (b).

\subsection{Gain from the first repetition}

The first labelled example provides the most information about how accurately each strategy can
recover the gain provided by full fine-tuning over the unadapted encoder
(Figure~\ref{fig:ch5-ranking-gain}(b); the complete strategy-by-budget matrix is given in
Table~\ref{tbl:ch5-results}(b) and Figure~\ref{fig:ch5-strategy-panel}(e)). In terms of recovery of gain from one labelled example prototypical
adaptation recovered 77\% of what full fine-tuning recovered on DB1, 63\% on DB2, and 63\% on
DB5, and it retained 78\%, 68\%, and 52\% of this gain at the largest budget (five labelled
examples) on DB1, four on DB2 and DB5 respectively. This was achieved without using any
optimiser, gradient steps, or changing any of the weights.

The linear probe is the least effective strategy among those tested. Using only labelled data to
refit the classifier head produced a macro F1 value of 0.169 on DB1 and 0.368 on DB2 at one
labelled example. Both values are lower than any method that allowed some movement in the
encoder, and on DB1 they are only slightly higher than the unadapted model. Therefore, the
representation learned during cross-user training is not linearly separable for a new user in
the way a picture of a frozen feature set plus a new classifier might suggest. To adaptively
personalize such encoders requires that either the representation adapts itself like the encoder
does during full-fine tuning, or that the decision boundary stops being a hyperplane and becomes
a distance to a class mean like the prototypes do. Freezing the encoder's representation and
maintaining a hyperplane as the decision boundary is the only combination that failed to produce
acceptable results from one labelled example.

\subsection{Full fine-tuning at larger budgets}

\begin{figure}[!htbp]\centering
\begin{subfigure}{\linewidth}\centering
\includegraphics[width=0.82\linewidth]{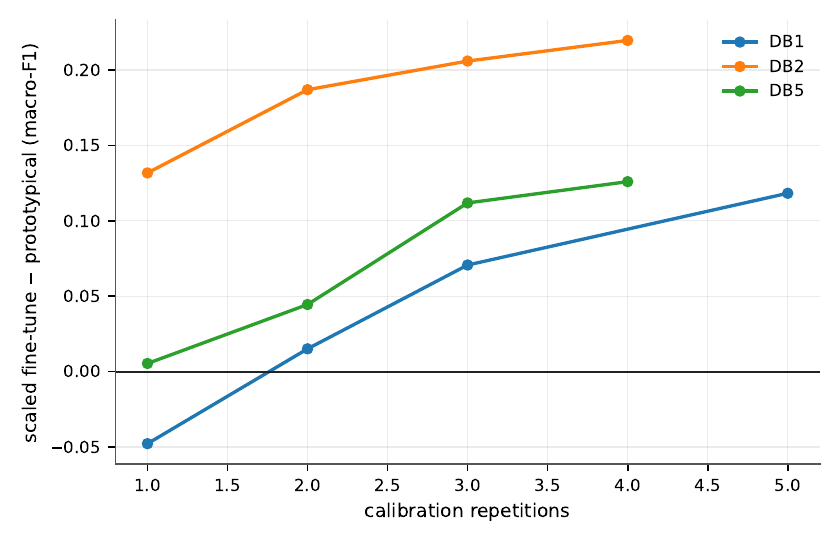}
\caption{}\label{fig:ch5-crossover}
\end{subfigure}

\vspace{1.5ex}

\begin{subfigure}{\linewidth}\centering
\includegraphics[width=0.95\linewidth]{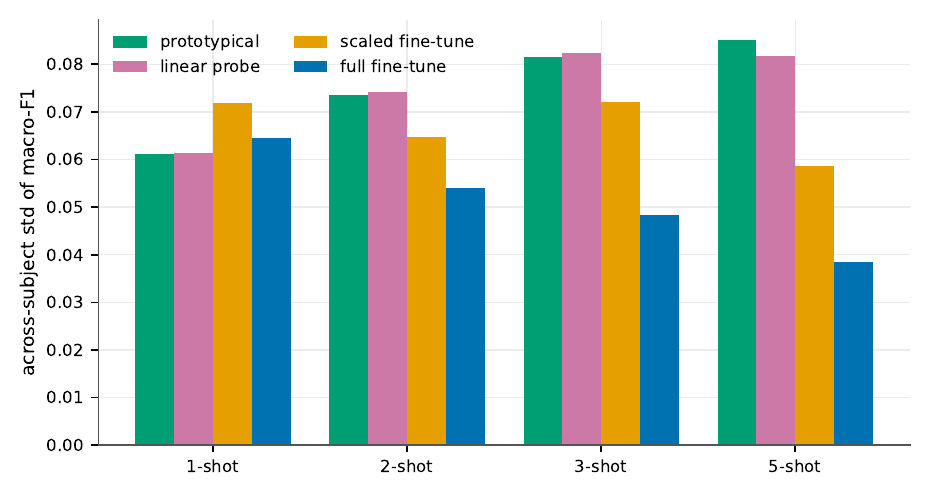}
\caption{}\label{fig:ch5-reliability}
\end{subfigure}
\caption{(a) Prototypical adaptation against the scaled fine-tune. Above zero the gradient-based method leads. The gradient-free rule is ahead only on DB1 at one repetition and level on DB5. (b) Across-subject standard deviation of each strategy at each calibration budget.}
\label{fig:ch5-crossover-reliability}
\end{figure}

Since full fine-tuning was the best performing strategy at every calibration budget on all three
databases, we focus our attention on comparing prototypical adaptation against the scaled
fine-tune. Prototypical adaptation changes no weights, and the scaled fine-tune is the other
method that moves the encoder; Table~\ref{tbl:ch5-comparisons}(a) and Fig.~\ref{fig:ch5-crossover-reliability}(a)
show that prototypical adaptation is better than the scaled fine-tune at one labelled example on
DB1 and is equal to it at one labelled example on DB5. By contrast, on DB2, where the signal is
richest and a gradient step is immediately useful, the scaled fine-tune is better than
prototypical adaptation starting at one labelled example. On DB1 and DB5, prototypical
adaptation falls behind the scaled fine-tune by around two to three labelled examples. The gap
between the gradient-free and gradient-based methods therefore narrows as the quality of signal
available decreases and the number of labels available decreases. This occurs in scenarios
similar to those experienced by prosthesis users when they initially put on their prosthetic
devices.

% Consolidated strategy-comparison table: the crossover table, the paired
% significance table and the reliability table, previously three separate
% floats, as (a), (b) and (c).
\begin{table}[!htbp]\centering
\caption{(a) Prototypical adaptation against the scaled fine-tune, the other strategy that moves the encoder. Full fine-tuning leads at every budget on all three databases, so it admits no crossing; against the scaled fine-tune the gradient-free rule is ahead only on DB1 at one repetition and level on DB5, and falls behind everywhere from two repetitions on. (b) Paired comparison of full fine-tuning against prototypical adaptation at each calibration budget. Positive differences favour full fine-tuning. (c) Across-subject standard deviation of each strategy, at the smallest and largest calibration budget. Lower is more reliable.}
\label{tbl:ch5-comparisons}
\begin{subtable}{\linewidth}\centering
\caption{}\label{tbl:ch5-crossover}
\begin{table}[t]\centering
\caption{Prototypical adaptation against the scaled fine-tune, the other strategy that moves the encoder. Full fine-tuning leads at every budget on all three databases, so it admits no crossing; against the scaled fine-tune the gradient-free rule is ahead only on DB1 at one repetition and level on DB5, and falls behind everywhere from two repetitions on.}\label{tbl:ch5-crossover}
{\small\setlength{\tabcolsep}{4pt}%
\begin{tabular}{lccc}
\toprule
Dataset & Proto.\ ahead up to & Scaled FT ahead from & Margin at one repetition \\
\midrule
DB1 & 1-shot & 2-shot & $+0.048$ \\
DB2 & none & 1-shot & $-0.132$ \\
DB5 & none & 1-shot & $-0.005$ \\
\bottomrule
\end{tabular}}
\end{table}

\end{subtable}

\vspace{1.5ex}

\begin{subtable}{\linewidth}\centering
\caption{}\label{tbl:ch5-significance}
{\small\begin{table}[t]\centering
\caption{Paired comparison of full fine-tuning against prototypical adaptation at each calibration budget. Positive differences favour full fine-tuning.}\label{tbl:ch5-significance}
\begin{tabular}{lcccc}
\toprule
Dataset & Budget & Mean $\Delta$ & Subjects favouring full & Wilcoxon $p$ \\
\midrule
DB1 & 1-shot & +0.095 & 27/27 & 1.5e-08 \\
DB1 & 2-shot & +0.147 & 27/27 & 1.5e-08 \\
DB1 & 3-shot & +0.160 & 27/27 & 1.5e-08 \\
DB1 & 5-shot & +0.178 & 27/27 & 1.5e-08 \\
DB2 & 1-shot & +0.221 & 40/40 & 1.8e-12 \\
DB2 & 2-shot & +0.248 & 40/40 & 1.8e-12 \\
DB2 & 3-shot & +0.246 & 40/40 & 1.8e-12 \\
DB2 & 4-shot & +0.248 & 40/40 & 1.8e-12 \\
DB5 & 1-shot & +0.064 & 9/10 & 3.9e-03 \\
DB5 & 2-shot & +0.135 & 10/10 & 2.0e-03 \\
DB5 & 3-shot & +0.210 & 10/10 & 2.0e-03 \\
DB5 & 4-shot & +0.232 & 10/10 & 2.0e-03 \\
\bottomrule
\end{tabular}
\end{table}
}
\end{subtable}

\vspace{1.5ex}

\begin{subtable}{\linewidth}\centering
\caption{}\label{tbl:ch5-reliability}
{\small\begin{table}[t]\centering
\caption{Across-subject standard deviation of each strategy, at the smallest and largest calibration budget. Lower is more reliable.}\label{tbl:ch5-reliability}
\begin{tabular}{lccccc}
\toprule
Dataset & Budget & Proto. & Linear probe & Scaled FT & Full FT \\
\midrule
DB1 & 1-shot & 0.061 & 0.061 & 0.072 & 0.065 \\
DB1 & 5-shot & 0.085 & 0.082 & 0.059 & 0.038 \\
DB2 & 1-shot & 0.091 & 0.107 & 0.090 & 0.065 \\
DB2 & 4-shot & 0.117 & 0.101 & 0.050 & 0.020 \\
DB5 & 1-shot & 0.055 & 0.043 & 0.059 & 0.064 \\
DB5 & 4-shot & 0.063 & 0.061 & 0.088 & 0.083 \\
\bottomrule
\end{tabular}
\end{table}
}
\end{subtable}
\end{table}

% Consolidated: the prototypical-against-scaled-fine-tune crossover plot and the
% across-subject standard-deviation plot, previously two separate floats.

\subsection{Per-subject behaviour}

\begin{figure}[!htbp]\centering
\begin{subfigure}{\linewidth}\centering
\includegraphics[width=0.96\linewidth]{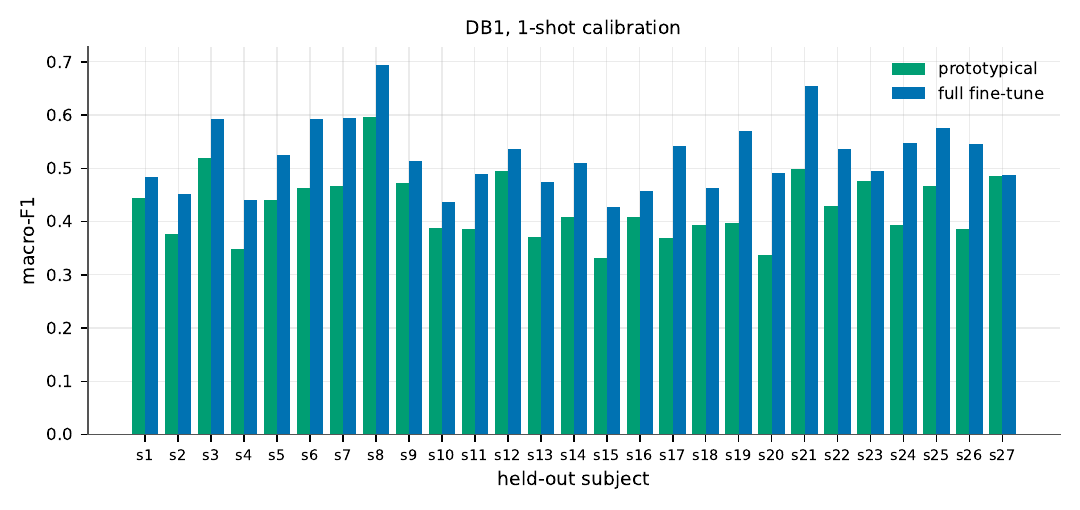}
\caption{}\label{fig:ch5-persubj-low}
\end{subfigure}

\vspace{1.5ex}

\begin{subfigure}{\linewidth}\centering
\includegraphics[width=0.96\linewidth]{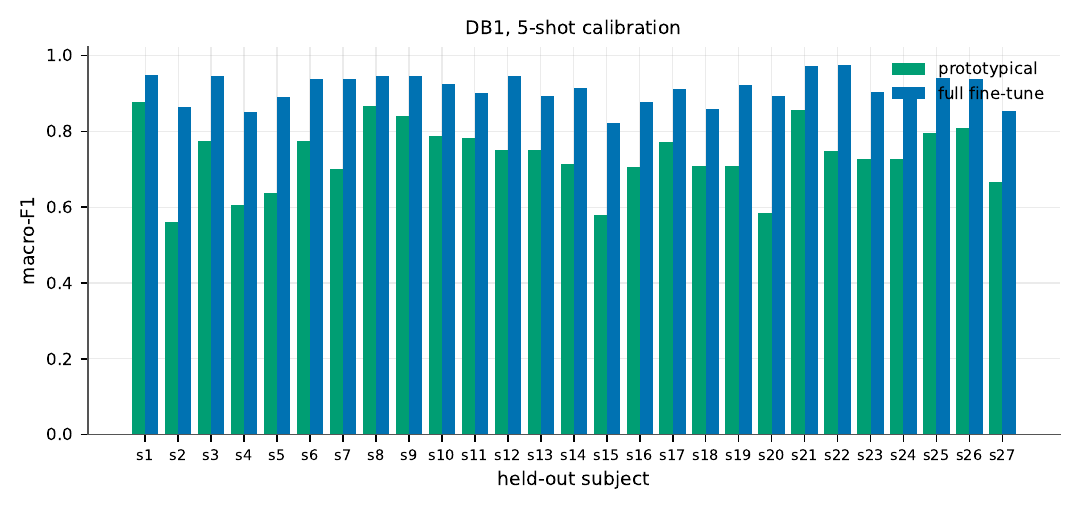}
\caption{}\label{fig:ch5-persubj-high}
\end{subfigure}
\caption{(a) Per-subject macro-F1 at the smallest calibration budget on DB1, prototypical against full fine-tuning. Full fine-tuning is ahead for all 27 held-out users. (b) Per-subject macro-F1 at the largest calibration budget on DB1. With more labelled repetitions full fine-tuning is ahead for every user.}
\label{fig:ch5-persubj}
\end{figure}

Table~\ref{tbl:ch5-persubject} and Figures~\ref{fig:ch5-persubj}(a)
and~\ref{fig:ch5-persubj}(b) provide a breakdown of how each strategy performed for each
subject individually. For all 27 subjects of DB1, full fine-tuning was better than prototypical
adaptation at both the smallest and largest budgets. Similarly, for all 40 subjects of DB2, full
fine-tuning was also better than prototypical adaptation. Additionally, for nine of ten subjects
at one labelled repetition and for all ten subjects at three labelled repetitions on DB5, full
fine-tuning was better than prototypical adaptation. The lone exception occurred on one subject
of DB5 at one labelled repetition, with a difference of only 0.007 macro-F1 units. There is thus
no group of subjects for whom prototypical adaptation is consistently more accurate than full
fine-tuning.

\begin{table}[t]\centering
\caption{Per-subject DB1 personalisation at the smallest and largest calibration budget: prototypical adaptation against full fine-tuning.}\label{tbl:ch5-persubject}
\begin{tabular}{lcccc}
\toprule
Subject & Proto. 1-shot & Full 1-shot & Proto. 5-shot & Full 5-shot \\
\midrule
s1 & 0.444 & 0.483 & 0.877 & 0.948 \\
s2 & 0.376 & 0.451 & 0.560 & 0.863 \\
s3 & 0.520 & 0.593 & 0.776 & 0.947 \\
s4 & 0.348 & 0.440 & 0.606 & 0.850 \\
s5 & 0.441 & 0.524 & 0.638 & 0.889 \\
s6 & 0.463 & 0.593 & 0.776 & 0.938 \\
s7 & 0.468 & 0.594 & 0.700 & 0.938 \\
s8 & 0.596 & 0.693 & 0.868 & 0.946 \\
s9 & 0.472 & 0.514 & 0.839 & 0.946 \\
s10 & 0.388 & 0.437 & 0.787 & 0.925 \\
s11 & 0.387 & 0.489 & 0.782 & 0.900 \\
s12 & 0.495 & 0.535 & 0.752 & 0.945 \\
s13 & 0.371 & 0.473 & 0.751 & 0.894 \\
s14 & 0.409 & 0.510 & 0.714 & 0.913 \\
s15 & 0.332 & 0.428 & 0.581 & 0.823 \\
s16 & 0.408 & 0.457 & 0.705 & 0.878 \\
s17 & 0.368 & 0.542 & 0.773 & 0.913 \\
s18 & 0.394 & 0.464 & 0.708 & 0.860 \\
s19 & 0.398 & 0.570 & 0.709 & 0.921 \\
s20 & 0.338 & 0.491 & 0.584 & 0.894 \\
s21 & 0.498 & 0.654 & 0.857 & 0.974 \\
s22 & 0.430 & 0.537 & 0.749 & 0.974 \\
s23 & 0.477 & 0.495 & 0.727 & 0.905 \\
s24 & 0.393 & 0.547 & 0.727 & 0.890 \\
s25 & 0.467 & 0.575 & 0.795 & 0.941 \\
s26 & 0.386 & 0.546 & 0.809 & 0.939 \\
s27 & 0.486 & 0.487 & 0.666 & 0.854 \\
\midrule
Mean & 0.428 & 0.523 & 0.734 & 0.911 \\
\bottomrule
\end{tabular}
\end{table}

% Consolidated: the per-subject DB1 bars at the smallest and at the largest
% calibration budget, previously two separate floats, as (a) and (b).

\subsection{Reliability across users}

In addition to accuracy comparison, we examine consistency across users through
Table~\ref{tbl:ch5-comparisons}(c) and Figure~\ref{fig:ch5-crossover-reliability}(b), which present the
across-subject standard deviation for each strategy. While there is no consistent winner in
terms of tightness of distribution across users, it is apparent that all distributions are
relatively wide, so individual variation in surface EMG remains substantial after
personalization.

\subsection{Accuracy against adaptation cost}

Finally, Table~\ref{tbl:ch5-cost-summary}(a) describes the various resource costs associated with deploying
each strategy. Given that prototypical adaptation uses only a single forward pass over the
calibration windows to compute class means in the encoder's embedding space and then ceases
activity entirely, i.e., it requires no optimiser state, no learning rate, no gradients, and no
per-user copies of model weights, it represents a very low-resource approach compared to the
other three strategies. Each of those must execute an optimizer run to learn model weights
appropriate to a particular user's calibration windows. Furthermore, as discussed previously,
the two fine-tuning variants additionally store personalized copies of all 3.3 million
parameters for every user.

\begin{figure}[!htbp]\centering
\begin{subfigure}{\linewidth}\centering
\includegraphics[width=0.82\linewidth]{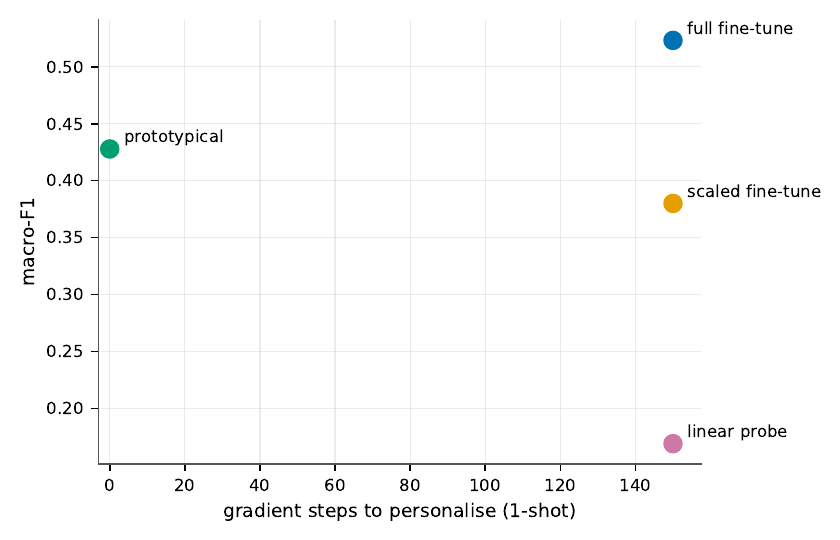}
\caption{}\label{fig:ch5-cost}
\end{subfigure}

\vspace{1.5ex}

\begin{subfigure}{\linewidth}\centering
\includegraphics[width=0.92\linewidth]{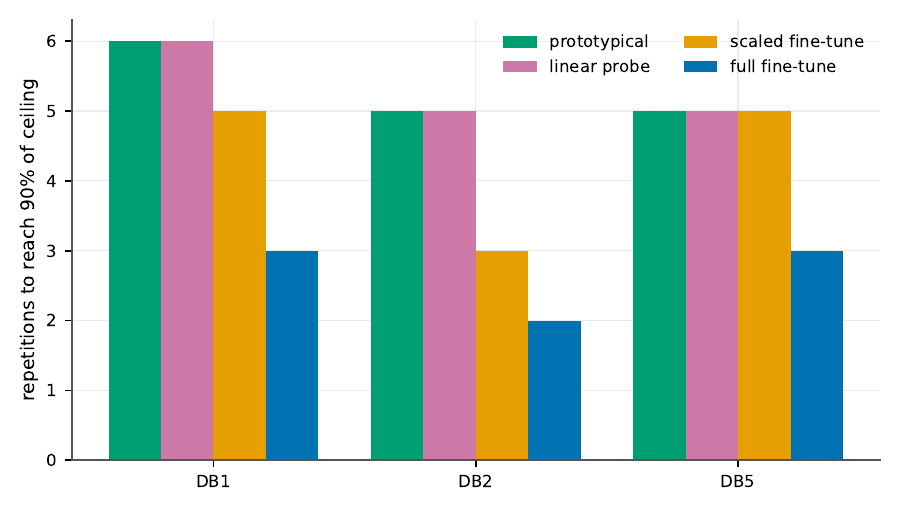}
\caption{}\label{fig:ch5-budget}
\end{subfigure}
\caption{(a) Accuracy against adaptation cost at one repetition on DB1. Prototypical adaptation needs no gradient steps and no parameter updates; the per-user LDA is cheaper still since it needs no encoder, and on DB1 prototypical adaptation is the more accurate of the two. (b) Calibration budget each strategy needs to reach ninety percent of the best score observed on the dataset. Lower is better.}
\label{fig:ch5-cost-panel}
\end{figure}

When viewed in light of these costs (Figure~\ref{fig:ch5-cost-panel}(a), Figure~\ref{fig:ch5-cost-panel}(b)) we
observe that prototypical adaptation achieves gains ranging from approximately half to roughly
four fifths that achievable via full-fine tuning at costs that are sufficiently low to be
executed on-device immediately upon donning a prosthetic device. Further, we note that
prototypical adaptation is less accurate than full-fine tuning regardless of how many labeled
examples were collected during calibration.

% Consolidated cost/baseline/summary table: the adaptation-cost table, the
% encoder-against-LDA table and the strategy summary table, previously three
% separate floats, as (a), (b) and (c).
\begin{table}[!htbp]\centering
\caption{(a) Adaptation cost of each personalisation strategy. Prototypical adaptation and the per-user LDA both need no optimisation; the LDA additionally needs no encoder. Cost is reported as optimisation requirement and parameters updated, not as measured latency. (b) Best encoder strategy, full fine-tuning at every budget on all three databases, against the per-user Hudgins and LDA pipeline at matched calibration budget. (c) Summary of personalisation strategies, ordered by adaptation cost from cheapest to most expensive. The ordering is not an accuracy ranking: full fine-tuning is the most accurate strategy at every calibration budget on all three databases.}
\label{tbl:ch5-cost-summary}
\begin{subtable}{\linewidth}\centering
\caption{}\label{tbl:ch5-cost}
{\footnotesize\begin{table}[t]\centering
\caption{Adaptation cost of each personalisation strategy. Prototypical adaptation and the per-user LDA both need no optimisation; the LDA additionally needs no encoder. Cost is reported as optimisation requirement and parameters updated, not as measured latency.}\label{tbl:ch5-cost}
\begin{tabular}{lll}
\toprule
Strategy & Optimisation & Parameters updated \\
\midrule
Prototypical & none & none \\
Linear probe & few epochs & head only ($<1\%$) \\
Scaled fine-tune & few epochs & all, encoder at $0.1\times$ LR \\
Full fine-tune & few epochs & all \\
Per-user LDA & none (closed form) & none (no encoder) \\
\bottomrule
\end{tabular}
\end{table}
}
\end{subtable}

\vspace{1ex}

\begin{subtable}{\linewidth}\centering
\caption{}\label{tbl:ch5-vslda}
{\footnotesize\begin{table}[t]\centering
\caption{Best encoder strategy, full fine-tuning at every budget on all three databases, against the per-user Hudgins and LDA pipeline at matched calibration budget.}\label{tbl:ch5-vslda}
\begin{tabular}{lcccc}
\toprule
Dataset & Budget & Encoder & LDA & $\Delta$ \\
\midrule
DB1 & 1-shot & 0.523 & 0.363 & +0.160 \\
DB1 & 3-shot & 0.830 & 0.593 & +0.238 \\
DB2 & 1-shot & 0.796 & 0.671 & +0.126 \\
DB2 & 3-shot & 0.967 & 0.857 & +0.110 \\
DB5 & 1-shot & 0.325 & 0.579 & -0.254 \\
DB5 & 3-shot & 0.588 & 0.802 & -0.214 \\
\bottomrule
\end{tabular}
\end{table}
}
\end{subtable}

\vspace{1ex}

\begin{subtable}{\linewidth}\centering
\caption{}\label{tbl:ch5-summary}
{\footnotesize\begin{table}[t]\centering
\caption{Summary of personalisation strategies, ordered by adaptation cost from cheapest to most expensive. The ordering is not an accuracy ranking: full fine-tuning is the most accurate strategy at every calibration budget on all three databases.}\label{tbl:ch5-summary}
\begin{tabular}{@{}l p{0.22\textwidth} p{0.44\textwidth}@{}}
\toprule
Strategy & Best regime & Outcome \\
\midrule
Prototypical & when no optimiser can run on the device & never the most accurate at any budget; recovers 52--78\% of full fine-tuning's gain over the unadapted encoder with no gradient step and no weight change \\
Linear probe & none identified & too weak alone; the encoder must move \\
Scaled fine-tune & when the cross-user prior must be preserved & second at every budget except DB1 at one repetition; never the best \\
Full fine-tune & every budget tested & most accurate on all three databases at every budget (27/27 DB1, 40/40 DB2, 9--10/10 DB5 subjects) \\
\bottomrule
\end{tabular}
\end{table}
}
\end{subtable}
\end{table}

% Consolidated: the accuracy-against-cost scatter and the budget-to-target bars,
% previously two separate floats, as (a) and (b).

\subsection{Comparison with the per-user linear classifier}

\begin{figure}[!htbp]\centering
\begin{subfigure}{\linewidth}\centering
\includegraphics[width=0.95\linewidth]{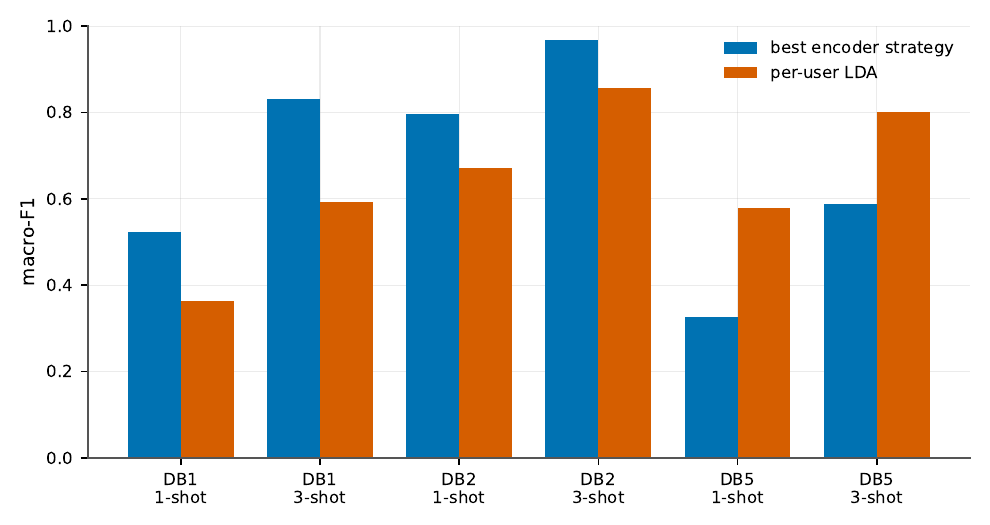}
\caption{}\label{fig:ch5-vslda}
\end{subfigure}

\vspace{1.5ex}

\begin{subfigure}{\linewidth}\centering
\includegraphics[width=0.82\linewidth]{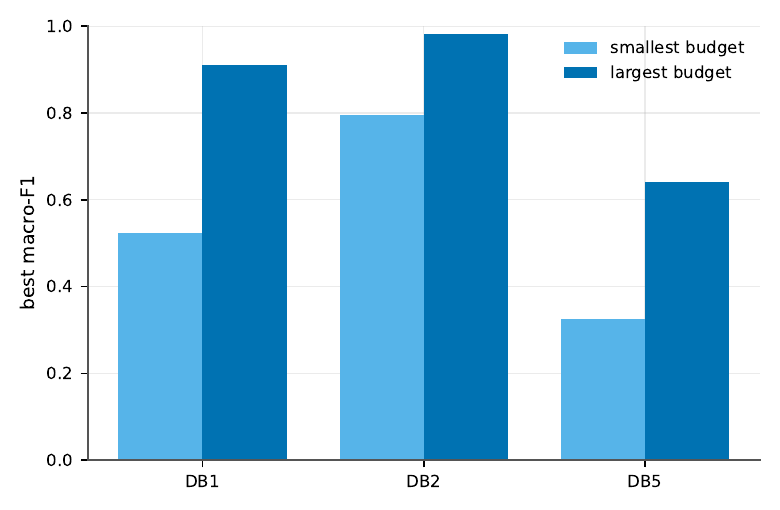}
\caption{}\label{fig:ch5-datasets}
\end{subfigure}
\caption{(a) Best encoder strategy against the per-user Hudgins and LDA pipeline at matched calibration budget, per dataset. (b) Headline personalisation result across the three databases: the best strategy at the smallest and largest calibration budget.}
\label{fig:ch5-lda-panel}
\end{figure}

Whether any of these results exceed the per-user linear classifier currently used by clinics
depends on which database is considered (Figure~\ref{fig:ch5-lda-panel}(b)). As predicted in earlier
work~\citep{odeyemi_encoder}
(Table~\ref{tbl:ch5-cost-summary}(b), Figure~\ref{fig:ch5-lda-panel}(a)), full-fine tuning is superior to the
Hudgins-LDA pipeline used by clinics for per-users on DB1 by 0.160 and 0.238 at one and three
labeled examples respectively, and by 0.126 and 0.110 for DB2. Conversely, full-fine tuning is
inferior to this baseline for per-users on DB5 by 0.254 and 0.214 at one and three labeled
examples respectively. Finally, prototypical adaptation is superior to this baseline for
per-users only on DB1 by 0.064 and 0.078 at one and three labeled examples respectively, and
inferior on DB2 by 0.095 and 0.136 and on DB5 by 0.318 and 0.424.

Again, this reflects the baseline strength account of the encoder
study~\citep{odeyemi_encoder}, albeit under a different question. We recall that the Hudgins-LDA
pipeline exhibits poor performance for per-users on DB1 due to low-quality signal samples and
consequently limited information contained in those signals. Thus, even prototypical adaptation,
which relies only on one forward pass through labeled data to compute class means, produces
superior results than this baseline for per-users on DB1. Conversely, since DB2 employs
high-frequency sampling (2000 Hz) resulting in a stronger signal than found in either of the
other two databases studied here, prototypical adaptation does not outperform this baseline for
per-users on DB2, and only the methods that move the encoder stay ahead of it.

Similarly to the encoder study~\citep{odeyemi_encoder}, however, cross-user training fails to
pay off on DB5, where every strategy falls below the baseline, because it includes only ten
subjects which is far too small for meaningful cross-user generalization.

% Consolidated: the encoder-against-LDA bars and the across-database headline
% bars, previously two separate floats, as (a) and (b).

\section{Discussion}

For accuracy alone, the answer is simply to fine-tune the entire model. This holds across all
calibration budgets, databases, and for virtually all subjects, and the paired margins against
the gradient-free option are significant throughout.

That the results have implications for engineering is the second. Prototypical adaptation which
computes class means in the embedding space of the encoder and alters nothing else, captures
roughly one-half to about four-fifths of the gain from full fine-tuning at a cost that is closer
to zero than to cheap. It uses no optimizer on the device and stores no weights per user. Given
a system must personalize immediately upon wear, and given that regime is like DB1's many
subjects and low signal fidelity, that is the strategy supported by this research. For regimes
such as those found on DB2 and DB5 where a per-user linear classifier is strong, that
closed-form baseline is even less expensive and more accurate, and remains the best
gradient-free choice. Therefore the personalization strategy used will depend on whether the
regime of use is like DB1 or more like DB2/DB5.

Design is constrained by the linear probe result. The widespread intuition that a good
representation only needs a fresh classifier is incorrect here. What distinguishes gestures from
a new user are not new hyperplanes over old features but either an adapted representation or a
distance based rule; a study that had only attempted linear probing would have incorrectly
concluded that the encoder is poor at transfer.

All results are off-line studies using intact limb participants, and were conducted within a
single recording session; they pertain to the cost of personalization and not to what occurs
after electrodes are removed and replaced, which a companion study treats
separately~\citep{odeyemi_intersession}. On DB5 none of the strategies exceed their respective
per-user baselines; this result is consistent with the mechanisms identified in the encoder
study~\citep{odeyemi_encoder}. Table~\ref{tbl:ch5-cost-summary}(c) summarises the four strategies and
their adaptation costs.

\section{Conclusion}

In terms of extracting accuracy from the same encoder, the same held out users, and the same
small number of labeled repetitions, full fine tuning clearly performs better than the other
three methods, and does so consistently enough that there is no exception among subsets of
subjects. The result which impacts how one might make an engineering decision however is that a
gradient free prototypical rule captures a large fraction of the benefit from fine-tuning while
leaving the model untouched; this allows for personalization to occur on a device during
wear-time instead of requiring a training run. How well each method may perform relative to a
per-user classifier that would be fitted by a clinic will depend on the specific database.

% ---------------------------------------------------------------------------
% Mandatory / recommended EAAI declarations. Order follows Elsevier house style;
% the generative-AI declaration is last because the guide requires it to be
% "placed in a new section before the references list".
% ---------------------------------------------------------------------------

\section*{CRediT authorship contribution statement}
\textbf{Jethro Odeyemi:} Conceptualization, Methodology, Software, Validation,
Formal analysis, Investigation, Data curation, Writing -- original draft,
Writing -- review \& editing, Visualization.

\textbf{W.J. (Chris) Zhang:} Conceptualization, Methodology, Resources, Supervision,
Project administration, Writing -- review \& editing.

\section*{Declaration of competing interest}
The authors declare no competing financial interests or personal relationships that could have
appeared to influence the work reported in this paper.

\section*{Funding}
This research did not receive any specific grant from funding agencies in the public, commercial, or not-for-profit sectors.

\section*{Ethics}
This work is a secondary analysis of previously published, publicly available data. No new
human-subject data were recorded for this study. Ethical approval and participant informed
consent for the original NinaPro recordings are reported by the database
authors~\citep{atzori2014electromyography,pizzolato2017comparison}.

\section*{Data availability}
This study uses only publicly available data. The three NinaPro databases analysed here (DB1,
DB2 and DB5) are distributed by the NinaPro consortium at \url{https://ninapro.hevs.ch/} and are
described by \citet{atzori2014electromyography} and \citet{pizzolato2017comparison}. No new data
were generated by this study.

\bibliographystyle{elsarticle-harv}
\bibliography{refs}

\end{document}